\documentclass{INTERSPEECH2023}

\title{Can We Trust Explainable AI Methods on ASR?  An Evaluation on Phoneme Recognition}

\newcommand{\ajitha}[1]{\textbf{\color{red}AR:{#1}}}

\newcommand{\xiaoliang}[1]{\textbf{\color{blue}Xiaoliang:{#1}}}
\name{Xiaoliang Wu, Peter Bell, Ajitha Rajan
}
\address{School of Informatics, University of Edinburgh}

\email{x.wu-53@sms.ed.ac.uk, peter.bell@ed.ac.uk, arajan@ed.ac.uk}
\usepackage{url}
\usepackage{color}
\usepackage{algorithm}
\usepackage{algpseudocode}
\usepackage{verbatim}

\usepackage{colortbl}
\usepackage{multirow}

\usepackage{xcolor}     
\usepackage{amsmath}
\usepackage{ulem}

\begin{document}
\renewcommand{\algorithmicrequire}{\textbf{Input:}}  
\renewcommand{\algorithmicensure}{\textbf{Output:}}
\maketitle
 
\begin{abstract}
Explainable AI (XAI) techniques have been widely used to help explain and understand the output of deep learning models in fields such as image classification and Natural Language Processing. Interest in using XAI techniques to explain deep learning-based automatic speech recognition (ASR) is emerging. 
but there is not enough evidence on whether these explanations can be trusted. 
To address this, we adapt a state-of-the-art XAI technique from the image classification domain, Local Interpretable Model-Agnostic Explanations (LIME),  to a model trained for a TIMIT-based phoneme recognition task. 
This simple task provides a controlled setting for evaluation while also providing expert annotated ground truth to assess the quality of explanations. We find a variant of LIME based on time partitioned audio segments, that we propose in this paper, produces the most reliable explanations, containing the ground truth 96\% of the time in its top three audio segments.  

\end{abstract}

\section{Introduction}
\label{sec:intro}
\vspace{-6pt}
 Explainable Artificial Intelligence (XAI) techniques have been proposed in recent years to help explain and understand the output of deep learning (DL) models used in image classification~\cite{LIME,DLIME,LIMEtree,k-LIME,deeplift,intergratedgradients,grad-cam} and NLP classification tasks~\cite{survey2018,survey2019,survey2021,surveyNLP2020,madsen2021post}. Our work focuses on post-hoc explanations, i.e., explanations an existing model that has been previously trained, which we treat as a black box system. Such post-hoc explanations are widely applicable as they can be used over models whose internal structure is not known. The need for explanations for AI systems has risen recently with the advent of safety regulations such as the recently proposed EU AI act\footnote{https://artificialintelligenceact.eu/the-act/} that requires explanations to help users better understand the decisions made by AI systems.  
 Although current explanation techniques have created a step change in providing reasons for predicted results from DL models, the question of whether the explanations themselves can be trusted has been largely ignored.  Some recent studies~\cite{wrongEP_amy,wrongEP_2,wrongEP_3} have demonstrated the limitations of current XAI techniques. For instance, \cite{wrongEP_amy} applied three different XAI techniques on a CNN-based breast cancer classification model and found the techniques disagreed on the input features used for the predicted output and in some cases picked background regions that did not include the breast or the tumour as  explanations. 

 Literature on evaluating the reliability of XAI techniques is still in its nascency and can be broadly divided into two branches - (1) Studies that assume the availability of expert annotated ground truth, maybe in the form of bounding boxes for images, to evaluate the accuracy of explanations~\cite{Lin_eva_gt,alignment_eva_1,alignment_eva_2,alignment_eva_3,eva_gt_yang,eva_gt_Arras,eva_gt_Adebayo,eva_gt_Holzinger} and (2) research that uses the idea of removing relevant (or important) features detected by an XAI method and verifying the accuracy degradation of the
retrained models \cite{hooker_performance_eva_7,petsiuk_performance_eva_3,Samek_eva_performance,yeh_performance_eva_6,eva_performance_ismail,arras2019eva_performance,lin_performance_eva_4,chen_performance_eva_2}. The first category requires human-annotated ground truth for evaluation while the second category incurs very high computational cost to verify accuracy degradation from retraining the models. 

Explanations for ASR models are only now beginning to emerge. \cite{ASR_EP} adapted image-based explanations for speech input in an ASR model. However, the quality and reliability of these explanations were not assessed. The primary obstacle to this in the ASR setting is the lack of  ground truth mapping from words in the transcription to the audio and this is challenging to obtain since the word outputs from the ASR may be driven by contextual prior knowledge from the language model not directly connected to the local speech input. 

In an effort to evaluate the reliability and trustworthiness of explanations in the ASR context, we use the TIMIT~\cite{TIMIT} Phoneme Recognition (PR) task using the standard receipe from the Kaldi toolkit~\cite{kaldi}, as a basic controllable task that is predictable with a phoneme language model and which provides a manual labelling and segmentation at the phomeme level. We then generate explanations for the PR system using LIME~\cite{LIME}, a popular XAI technique developed for images that uses an interpretable linear regression model to locally approximate the prediction of the target black box DL model.  LIME only works for classification tasks, so it is necessary to make a number of changes.
 First, we classify every phoneme in a PR transcription as correct or incorrect based on comparison with the expected transcription.  Second, we apply LIME to generate explanations for each phoneme in the transcription using input speech perturbations. Third, we improve performance of LIME by restricting perturbations of the input audio to be within a limited window around the phoneme of interest using two LIME variations, LIME Window Segment\footnote{In our paper, segments of audio refer to distinct sections audio obtained from manual segmentation of the TIMIT dataset or timestamps.} (LIME-WS) and LIME Time Segment (LIME-TS).
 We evaluate reliability of the basic LIME explanations and the variants, LIME-WS and LIME-TS, for the TIMIT PR task on Kaldi using the ground truth labelling of the TIMIT dataset. We found explanations with LIME-TS are the most reliable for the TIMIT Phoneme Recognition task using Kaldi, capturing the ground truth 96\% of the time in the top three segments of the explanation.  LIME-TS outperforms LIME and LIME-WS by up to 44.8\% and 17\% on \texttt{All} speakers, respectively.

Source code for \texttt{X-PR} and examples are available at  \url{https://anonymous.4open.science/r/X-PR-4560}.

\section{Methodology}
\label{sec: methodlogy}
\vspace{-6pt}

\begin{figure}[ht]
    \centering
    \includegraphics[width = \linewidth]{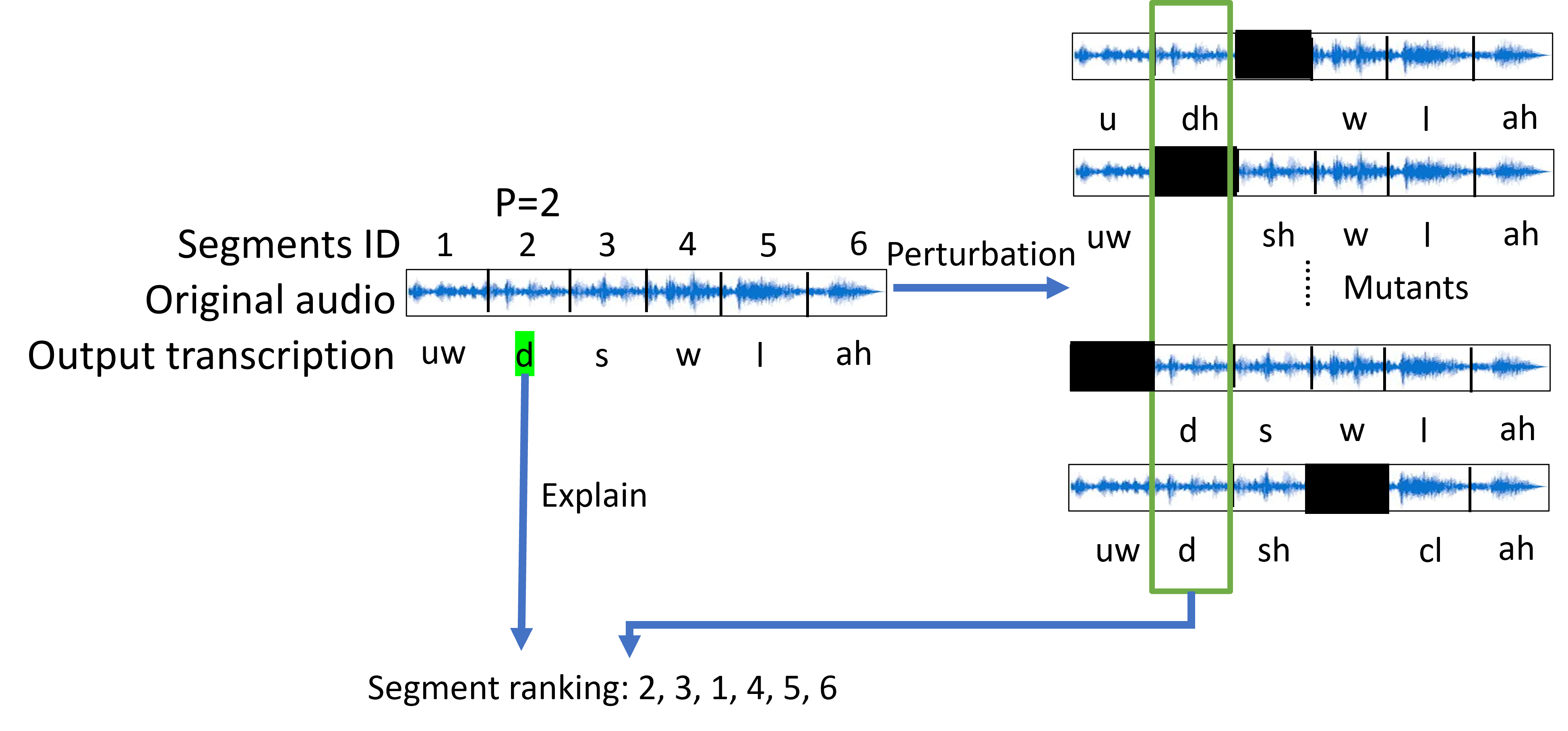}
    \caption{An outline of generating an explanation for a phoneme appearing in the output transcription.}
    \label{fig: explanation}
\end{figure}

\begin{figure}[ht]
    \centering
    \includegraphics[width = \linewidth]{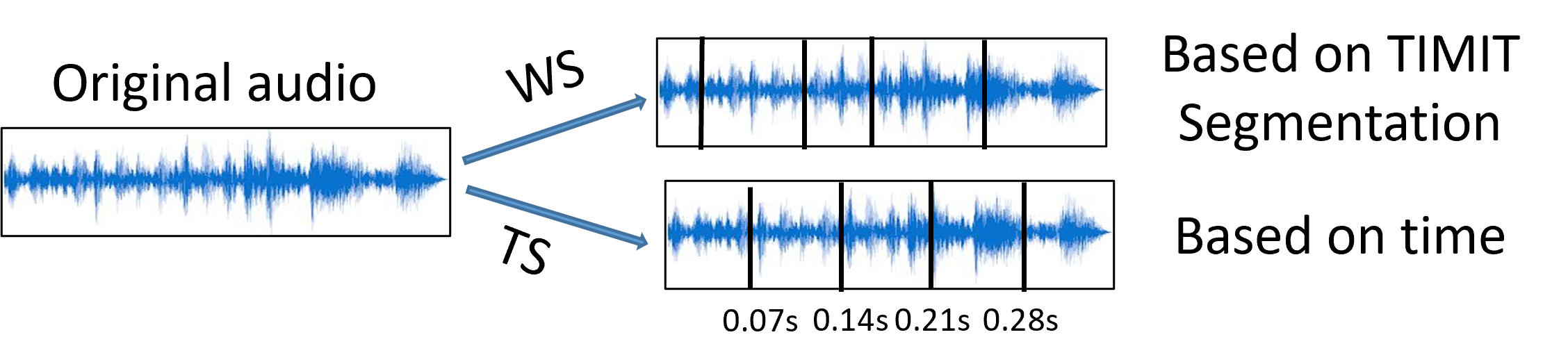}
    \caption{Different segmentation used by (LIME-WS,LIME) and LIME-TS.}
    \label{fig: segmentation}
\end{figure}

Figure~\ref{fig: explanation} presents a high level overview of generating an explanation for a phoneme in the transcription output. We start with an input audio (divided into segments available in TIMIT). We then perturb the audio by masking out segments (shown as blacked out segments where the sample points are set to zero) randomly to generate several perturbed audio, referred to as mutants. For the phoneme of interest, shown as `\texttt{d}' in Figure~\ref{fig: explanation}, we compute the importance ranking of segments in the audio as an explanation for \texttt{d} which is based on the effect of the perturbations on the phoneme output. 
 
We discuss the LIME explanation method and its variants, LIME-WS and LIME-TS,  to generate importance ranking of segments for each phoneme in more detail later in this section. To apply the LIME technique, we first need to treat the PR task as a classification task. To do this, we attach 0 or 1 label to every phoneme in the output transcription by aligning and comparing it with the expected transcription that is available in the TIMIT dataset. We implement classification of each phoneme output with the NIST sclite scoring tool\footnote{\url{https://github.com/usnistgov/SCTK}}.

\subsection{Explanations using LIME and its variants for PR}

In this section, we start by describing the base case which is a straight forward adaptation of LIME to work on the PR task. We then describe our variants, LIME-WS and LIME-TS, that applies perturbations to segments within a fixed window. 

\noindent \textbf{Base LIME Explanations:}
LIME, proposed in ~\cite{LIME}, is an XAI technique that is black-box, allowing it to be applied to any model without requiring information on its structure. 
Given a complex neural network (NN)  model $f(x)$ that takes in an input $x$ and produces an output $y$, the goal of LIME is to mimic the behavior of a complex model $f(x)$ with a weighted linear regression model $g(x)$ in the local area of a specific instance of interest $x$. The weighted linear regression model $g(x)$ is defined as:
\begin{equation}
g(x) = w_0 + w_1 x_1 + w_2 x_2 + ... + w_d x_d
\label{Eq:gx}
\end{equation}
where $w_0$ is the intercept term, $w_1$ to $w_d$ are the weights assigned to each feature, and $x_1$ to $x_d$ are the feature values of the instance $x$. Among them, $w_1$ to $w_d$ denote what we call the \textit{importance score} of each feature.

For the PR task, the complex NN model $f(x)$ is the Kaldi system. The specific instance $x$ is the original audio. To fit $g(x)$ and get $w_1$ to $w_d$, LIME needs several perturbed instances of $x$ and their outputs. As shown in the right side of the Figure~\ref{fig: explanation}, mutants of the original audio $x$ are created by masking out segments randomly. Input features $x$ are the segments.  
Values of features in the perturbed audio are 1 or 0, where 1 means that the segment at this position has not been masked while 0 implies the segment is masked. For example, for the mutant in the upper right corner of Figure~\ref{fig: explanation}, the third segment is masked. 

 For each mutant $m_i$, we align its transcription against the original (unmasked) output transcription, $y$. After alignment, we may find some phonemes match the original output transcription while some others are incorrect. For those correct phonemes, output of $m_i$ - represented as $f^p(m_i)$ - will be 1 while for others, $f^p(m_i)$ will be 0. $p$ refers to the position of the chosen phoneme. $f(m_i)$ is a sequence of phonemes, but $f^p(m_i)$ is a binary label to indicate the existence or not of the chosen phoneme in $f(m_i)$.
We see in Figure~\ref{fig: explanation}, we want to generate the explanation for phoneme \texttt{d} that is highlighted in the output transcription. The $p$ of \texttt{d} is 2 which is shown on top of that. the LIME model will compute how the masked segment in each of the mutants affects the output phoneme \texttt{d} (bounded with a green box).   If the masked segment changes the output phoneme \texttt{d} at that position, then it will have a high ranking (aggregated over many mutants with masked segments). 

Using the mutants and the associated binary label after alignment to original transcription, LIME will start to fit $g(x)$ using the locally weighted least squares objective function, which is defined as:
\begin{equation}
L(g)_p = \sum_{i=1}^n we_i (f^p(m_i) - g(m_i))^2
\end{equation}
In this equation, $n$ is the number of mutants and $we_i$ is a weight assigned to each mutant $m_i$ that reflects its closeness to the audio of interest $x$. It is computed as the cosine similarity between the instance $x$ and the mutant $m_i$, which is:
\begin{equation}
we_i = Cosine\_Similarity(x, m_i)
\end{equation}

The weights $w_1$ to $w_d$ in the fitted linear regression model, $g(x)$, indicate the importance score of different audio segments for the selected output phoneme. We treat the ranking of segments based on their importance score as the explanation for each phoneme, as shown in Figure~\ref{fig: explanation}.

\textbf{Segment-based LIME with a sliding window(LIME-WS):} 
As seen in Base LIME, mutants for the original audio $x$ are created by masking random segments. However, owing to the local nature of the PR task, the output phoneme being explained is not likely to be influenced by distant audio segments.  
Taking this into account, will eliminate worthless mutants where the masked out audio segments have no effect on the phoneme being examined, making the computation more effective. 

To realize this idea, we use a fixed length sliding window during the generation of perturbations for LIME explanations.
The sliding window slides from left to right one segment at a time. 
Within the range delimited by the sliding window, a pre-determined number of segments are randomly chosen for masking, while keeping the segments outside the sliding window unchanged. We hypothesize that focussing on perturbations within this sliding window will result in higher qualtiy explanations. Other steps in LIME-WS for fitting the linear regression model remain the same as LIME.  

\textbf{Time Segment-based LIME with a sliding window(LIME-TS):} Audio segmentation in LIME and LIME-WS is from the TIMIT dataset, wherein linguistic experts analyze the audio and partition it into distinct segments. 
However, such manually generated segmentations may not be readily available in other datasets (such as Librispeech and Common voice\cite{commonvoice:2020}). To overcome this challenge and generalize the applicability of our segment-based explanation technique,  we investigate an approach for dividing the original audio into non-overlapping segments of equal duration using timestamps. Figure~\ref{fig: segmentation} shows the original audio segmented in two different ways - one using the existing segments in TIMIT and the other using time. We use the time segments as a replacement for the segments in LIME-WS and LIME. Among the 630 audio files we use in the TIMIT dataset, the average duration of TIMIT segments is 78ms. We choose a time segment duration of 70ms to remain comparable to TIMIT segments and to accommodate TIMIT audio with slightly smaller segment lengths. 

\section{Experiments}
\vspace{-6pt}
We assess the reliability of explanation techniques using the  TIMIT Phoneme Recognition model from Kaldi, trained with default settings. To generating explanations, we use the TIMIT dataset too: TIMIT is chosen since it provides ground truth mapping from phonemes in the transcription to the input speech. Moreover, each sentence includes information such as the speaker's gender, dialect type, which facilitates our further analysis.
Across the entire TIMIT dataset, we select all 630 speakers for explanation generation. For each speaker, we generate explanations from one sentence `SA1' sentence that they have in common. Having a common sentence across speakers enables us to compare potential effects of factors like gender on reliability of explanations. 
We evaluate all three explanation techniques mentioned in Section~\ref{sec: methodlogy}, namely, LIME, LIME-WS and LIME-TS using the validity metric described in the next section.


{\textbf{Validity Metric}}
The validity metric evaluates the reliability of explanation methods quantitatively. Specifically, for every audio in the TIMIT dataset, each phoneme in its transcription is associated with a ground truth segment in the audio  (manually annotated by linguists). An explanation for a given phoneme will produce an importance ranking of the segments in the original audio. We assess the explanation by checking if the ground truth segment that it is associated with is in the top 1, top 5 or top 10 segments in the explanation's importance ranking. We have a validity metric for each of these.  
More specifically, we define $validity_1$=$\frac{N_1}{N}$, in which $N$ represents the total number of phonemes in the dataset, while $N_1$ refers to the number of phonemes whose  ground truth segment appears as the top ranked segment in the corresponding explanation. Similarly, we define $validity_3$=$\frac{N_3}{N}$ and $validity_5$=$\frac{N_5}{N}$. $N_3$ and $N_5$ refer to the number of phonemes whose ground truth segment appears in the top 3 or 5 ranks in the corresponding explanation, respectively.  
All three validity metrics are in the range of $0$ to $1$ with a higher value indicating better reliability. As is expected, higher values of $validity_1$ is harder to achieve than $validity_3$ which is in turn harder than $validity_5$.  

\section{Results and Analysis}
\label{sec: results}

\begin{table*}[h]
\centering
\begin{tabular}{|c|
>{\columncolor[HTML]{FFCCC9}}c 
>{\columncolor[HTML]{FFCCC9}}c 
>{\columncolor[HTML]{FFCCC9}}c |
>{\columncolor[HTML]{FFFFC7}}c 
>{\columncolor[HTML]{FFFFC7}}c 
>{\columncolor[HTML]{FFFFC7}}c |
>{\columncolor[HTML]{ECF4FF}}c 
>{\columncolor[HTML]{ECF4FF}}c 
>{\columncolor[HTML]{ECF4FF}}c |}
\hline
                   & \multicolumn{3}{c|}{\cellcolor[HTML]{FFCCC9}LIME / Random ranking}                                                                                                  & \multicolumn{3}{c|}{\cellcolor[HTML]{FFFFC7}LIME-WS / Random Ranking}                                                                                               & \multicolumn{3}{c|}{\cellcolor[HTML]{ECF4FF}LIME-TS / Random Ranking}                                                                                                                     \\ \cline{2-10} 
\multirow{-2}{*}{} & \multicolumn{1}{c|}{\cellcolor[HTML]{FFCCC9}All}       & \multicolumn{1}{c|}{\cellcolor[HTML]{FFCCC9}Female}    & Male                             & \multicolumn{1}{c|}{\cellcolor[HTML]{FFFFC7}All}       & \multicolumn{1}{c|}{\cellcolor[HTML]{FFFFC7}Female}    & Male                             & \multicolumn{1}{c|}{\cellcolor[HTML]{ECF4FF}All}                              & \multicolumn{1}{c|}{\cellcolor[HTML]{ECF4FF}Female}    & {\color[HTML]{333333} Male}      \\ \hline
$validity_1$        & \multicolumn{1}{c|}{\cellcolor[HTML]{FFCCC9}0.40/0.0}  & \multicolumn{1}{c|}{\cellcolor[HTML]{FFCCC9}0.35/0.0}  & {\color[HTML]{3166FF} 0.42/0.0}  & \multicolumn{1}{c|}{\cellcolor[HTML]{FFFFC7}0.49/0.03} & \multicolumn{1}{c|}{\cellcolor[HTML]{FFFFC7}0.43/0.03} & {\color[HTML]{3166FF} 0.50/0.03} & \multicolumn{1}{c|}{\cellcolor[HTML]{ECF4FF}{\color[HTML]{FE0000} 0.86/0.03}} & \multicolumn{1}{c|}{\cellcolor[HTML]{ECF4FF}0.84/0.02} & {\color[HTML]{3166FF} 0.86/0.03} \\ \hline
$validity_3$        & \multicolumn{1}{c|}{\cellcolor[HTML]{FFCCC9}0.62/0.06} & \multicolumn{1}{c|}{\cellcolor[HTML]{FFCCC9}0.54/0.06} & {\color[HTML]{3166FF} 0.62/0.06} & \multicolumn{1}{c|}{\cellcolor[HTML]{FFFFC7}0.76/0.09} & \multicolumn{1}{c|}{\cellcolor[HTML]{FFFFC7}0.72/0.09}  & {\color[HTML]{3166FF} 0.77/0.09} & \multicolumn{1}{c|}{\cellcolor[HTML]{ECF4FF}{\color[HTML]{FE0000} 0.96/0.12}} & \multicolumn{1}{c|}{\cellcolor[HTML]{ECF4FF}0.94/0.11} & {\color[HTML]{3166FF} 0.96/0.13} \\ \hline
 $validity_5$        & \multicolumn{1}{c|}{\cellcolor[HTML]{FFCCC9}0.67/0.13} & \multicolumn{1}{c|}{\cellcolor[HTML]{FFCCC9}0.59/0.12} & {\color[HTML]{3166FF} 0.67/0.13} & \multicolumn{1}{c|}{\cellcolor[HTML]{FFFFC7}0.83/0.15} & \multicolumn{1}{c|}{\cellcolor[HTML]{FFFFC7}0.81/0.14} & {\color[HTML]{3166FF} 0.84/0.15} & \multicolumn{1}{c|}{\cellcolor[HTML]{ECF4FF}{\color[HTML]{FE0000} 0.97/0.16}} & \multicolumn{1}{c|}{\cellcolor[HTML]{ECF4FF}0.96/0.15} & {\color[HTML]{3166FF} 0.97/0.16} \\ \hline
\end{tabular}
\caption{$validity_1$, $validity_3$ and $validity_5$ of explanations generated by three explanation methods and randomly sorted method (Right side of every slash) across gender using Kaldi PR.}
\label{tab: all}
\end{table*}

\begin{figure}[ht]
    \centering
    \includegraphics[width = \linewidth]{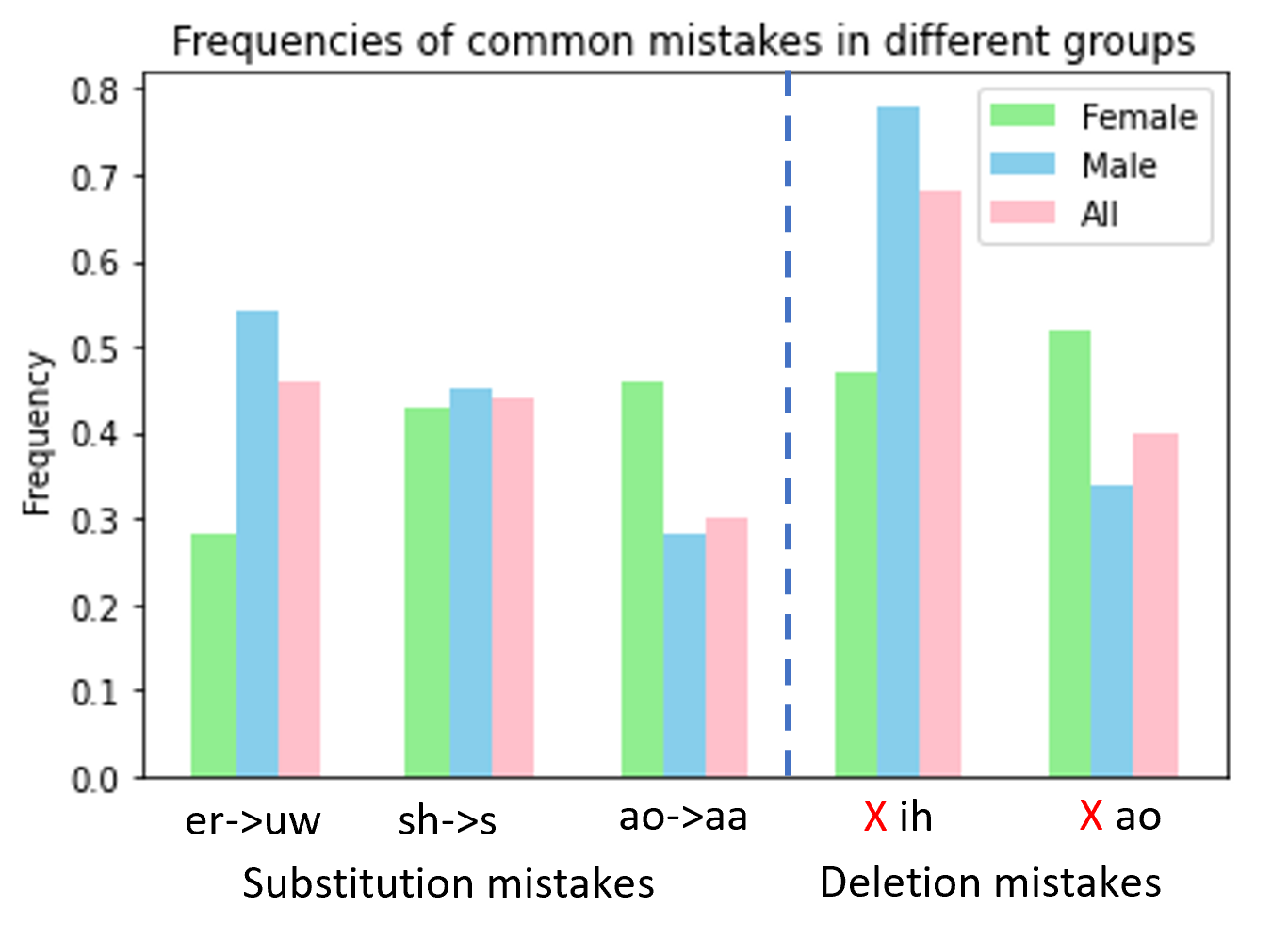}
    \caption{The top five most frequently occurring transcription mistakes and their corresponding frequencies on different groups. There are three substitution mistakes on the left of the dashed blue line and two deletion mistakes on the right. For example, $er\rightarrow uw$ means that er is replaced by uw and $\textcolor{red}{X} ih$ means that ih is deleted.}
    \label{fig: mistake}
\end{figure}

\begin{table}[h]
\centering
 \resizebox{\columnwidth}{15mm}{
\begin{tabular}{|
>{\columncolor[HTML]{FFCCC9}}c 
>{\columncolor[HTML]{FFCCC9}}l |
>{\columncolor[HTML]{FFCCC9}}c 
>{\columncolor[HTML]{FFCCC9}}l |
>{\columncolor[HTML]{FFCCC9}}c 
>{\columncolor[HTML]{FFCCC9}}l |}
\hline
\multicolumn{2}{|c|}{\cellcolor[HTML]{FFCCC9}All}                                                                          & \multicolumn{2}{c|}{\cellcolor[HTML]{FFCCC9}{\color[HTML]{000000} Female}}                                                & \multicolumn{2}{c|}{\cellcolor[HTML]{FFCCC9}{\color[HTML]{000000} Male}}                                                  \\ \hline
\multicolumn{2}{|l|}{\cellcolor[HTML]{FFCCC9}\begin{tabular}[c]{@{}l@{}}Segment Position \\ (Phoneme output)\end{tabular}} & \multicolumn{2}{l|}{\cellcolor[HTML]{FFCCC9}\begin{tabular}[c]{@{}l@{}}Segment Position \\ (Phoneme output)\end{tabular}} & \multicolumn{2}{l|}{\cellcolor[HTML]{FFCCC9}\begin{tabular}[c]{@{}l@{}}Segment Position \\ (Phoneme output)\end{tabular}} \\ \hline
\multicolumn{2}{|c|}{\cellcolor[HTML]{FFCCC9}8 (d)}                                                                        & \multicolumn{2}{c|}{\cellcolor[HTML]{FFCCC9}{\color[HTML]{000000} 8 (d)}}                                                 & \multicolumn{2}{c|}{\cellcolor[HTML]{FFCCC9}{\color[HTML]{000000} 34 (er)}}                                               \\ \hline
\multicolumn{2}{|c|}{\cellcolor[HTML]{FFCCC9}9 (aa)}                                                                       & \multicolumn{2}{c|}{\cellcolor[HTML]{FFCCC9}{\color[HTML]{000000} 9 (aa)}}                                                & \multicolumn{2}{c|}{\cellcolor[HTML]{FFCCC9}{\color[HTML]{000000} 8 (d)}}                                                 \\ \hline
\multicolumn{2}{|c|}{\cellcolor[HTML]{FFCCC9}34 (er)}                                                                      & \multicolumn{2}{c|}{\cellcolor[HTML]{FFCCC9}{\color[HTML]{000000} 7 (dcl)}}                                               & \multicolumn{2}{c|}{\cellcolor[HTML]{FFCCC9}{\color[HTML]{000000} 32 (y)}}                                                \\ \hline
\end{tabular}}
\caption{The top three important segments in LIME-WS explanation for the $er\rightarrow uw$ mistake and the corresponding phoneme outputs in paranthesis with speaker groups \texttt{All}, \texttt{Female} and \texttt{Male}.}

\label{tab: behaviours_explanation}
\end{table}

We present the reliability of the explanation techniques in terms of the validity metrics.
\subsection{Comparison of Explanation Techniques.}
All three techniques, LIME, LIME-WS, LIME-TS, are able to generate explanations for every input audio in our dataset.
Table~\ref{tab: all} shows $validity_1$, $validity_3$ and $validity_5$ of explanations generated by the three explanation techniques across \texttt{All} speakers, and over the \texttt{Male} and \texttt{Female} speakers separately. 

As a baseline for comparing the importance ranking generated by the different explanations, we randomly sort the segments to produce a random ranking and compute $validity_1$, $validity_3$ and $validity_5$ using it. We show the validity scores with random ranking in  Table~\ref{tab: all} as the value following the $/$ symbol in each cell. 
We find all three explanation techniques are significantly more reliable than random ranking across all three validity metrics. For instance, as seen in Table~\ref{tab: all}, the performance of LIME-WS on the three validity metrics across all speakers is $0.49$, $0.76$, and $0.83$, which is substantially higher than the corresponding values of $0.03$, $0.10$, and $0.15$ achieved by random ranking. The other two explanation techniques, LIME and LIME-TS, also perform better than random ranking. (Statistically significant difference was confirmed with one-way Anova followed by post-hoc Tukey's test~\cite{tukey}). 

Based on the results presented in Table~\ref{tab: all}, we observe that both LIME-WS and LIME-TS outperform LIME across all three metrics and speaker groups. For instance, LIME-WS and LIME-TS outperformed LIME by 24\% and 44.8\%  on $validity_5$ with \texttt{All} speakers. 
This outcome is in line with our expectations, as the sliding window mechanism introduced in LIME-WS and LIME-TS ensures that perturbations are restricted to a local range. Specifically, in the context of PR tasks, the influence of a phoneme is typically confined to a small number of adjacent phonemes. Therefore, when generating an explanation for a phoneme, mutations clustered around it enable the explanation methods to more accurately pinpoint the most relevant audio segments. 

For LIME-TS versus LIME-WS, we find LIME-TS performs better than LIME-WS in all cases. For example, LIME-TS outperforms LIME-WS by 17\% on $validity_5$ over \texttt{All} speakers. We believe this is because when we use fixed-length time segments in LIME-TS rather than TIMIT defined segments in the LIME-WS, nearly 40\% of ground truth audio segments (that are TIMIT based) overlap with multiple time segments (usually 2). In such cases, multiple time segments may be ranked with high importance. When we compute validity, we check if \emph{at least one} of the overlapping time segments is present in top 1, 2 or 5 ranked segments (depending on the validity metric). We believe the overlap along with the at least one criteria makes it easier for LIME-TS explanations to achieve higher validity scores.  

Overall, we find LIME-TS to be most reliable among the three explanation techniques, capturing the ground truth in 96\% or 97\% of the cases when considering the top 3 or 5 segments in the explanations, respectively. Additionally, it is more easily generalizable than LIME-WS owing to its use of time in segment definition rather than an expert annotated audio specific segment definition that may not be widely available. 



\subsection{\texttt{Male} versus \texttt{Female} Speakers.}
Table~\ref{tab: all} reveals that, in terms of gender,  the validity metrics for all three explanation techniques are higher for \texttt{Males} than  \texttt{Females}. This finding is not surprising given that the models were trained on the TIMIT dataset which has 70\% male speakers and only 30\% female speakers. As a result, Kaldi is more adept at capturing the characteristics of male utterances, even when both male and female speakers are uttering the same sentence. Consequently, our explanation techniques are more effective at identifying the ground truth segment for \texttt{Males}.


We explored the most commonly occurring transcription errors in the different speaker groups. 
Figure~\ref{fig: mistake} illustrates the five most commonly occurring transcription errors and their corresponding frequencies across the three speaker groups: \texttt{All, Female}, and \texttt{Male}. We find that for four of the five mistakes, except for the substitution error, $sh\rightarrow s$, there is a  significant difference in the frequency of the mistakes between \texttt{Female} and \texttt{Male} speakers. For instance, the probability of $er\rightarrow uw$ mis-transcription is higher among \texttt{Male} speakers (0.54) compared to \texttt{Female} speakers (0.28). 
We conducted a Wilcoxon Signed Rank Test at 5\% significance level and confirmed that the differences in error frequencies across all mistakes between the \texttt{Males} and \texttt{Females} is statistically significant.   

Explanations can help investigate possible causes for difference in error frequencies between genders. We illustrate this with one of the top five transcription mistakes depicted in Figure~\ref{fig: mistake}, namely $er\rightarrow uw$ substitution error.
We compare the audio segments from explanations generated with the LIME-WS  technique  between \texttt{Malse} and \texttt{Females} for this phoneme error. 
We identify the top three segments in explanations with the highest frequencies for this error and record their corresponding segment IDs. The results shown in Table~\ref{tab: behaviours_explanation}, reveal differences 
 in the top three most frequent segments in explanations across speaker groups. For \texttt{Female} speakers, all three of the top segments are located around position 8, whereas, for \texttt{Male} speakers, two of the top three segments are situated around position 34. On further inspection, we find the $er\rightarrow uw$ substitution error occurs in two distinct locations in sentence SA1 -- one occurs at position 6 where $er$ is replaced by $uw$ and is followed by phonemes $vcl\ d\ aa$, while the second instance occurs at position 34, which is surrounded by phoneme $y$. We find among \texttt{Female} speakers, 64\% of the $er\rightarrow uw$ mistakes are associated with position 6 (versus 47\% for \texttt{Male}). The substitution mistake is more evenly distributed between both positions for \texttt{Male} speakers. This information can be used to improve the model with further training for \texttt{Female} and \texttt{Male} speakers. We, thus, find explanations are a useful tool to examine errors in transcriptions and compare speaker groups.

\section{Conclusion}

In this study, we adapt LIME, originally designed for the image domain, to generate explanations for the TIMIT Phoneme Recognition (PR) task. To better suit the characteristics of PR and facilitate generalization to other datasets, we propose two LIME-based explanation techniques, namely LIME-WS and LIME-TS. To assess the reliability of these techniques, we conduct experiments under controlled conditions with ground truth available. Our findings indicate that LIME-TS is the most reliable, containing the ground truth audio segment in phoneme output explanations in 96\% of the cases (considering top three audio segments in the explanaitons).


While our evaluation of explanation technique reliability on the TIMIT PR task provides initial insights, it remains a significant challenge to quantitatively assess these techniques on more complex speech tasks owing to the involvement of other components, such as a pre-trained language model, and the presence of long-span dependencies that make it challenging to create a ground truth mapping from word outputs to audio segments. We aim to address evaluation of explanations over more complex speech tasks in the future.

\bibliographystyle{IEEEtran}
\bibliography{mybib}

\begin{thebibliography}{10}
\providecommand{\url}[1]{#1}
\csname url@samestyle\endcsname
\providecommand{\newblock}{\relax}
\providecommand{\bibinfo}[2]{#2}
\providecommand{\BIBentrySTDinterwordspacing}{\spaceskip=0pt\relax}
\providecommand{\BIBentryALTinterwordstretchfactor}{4}
\providecommand{\BIBentryALTinterwordspacing}{\spaceskip=\fontdimen2\font plus
\BIBentryALTinterwordstretchfactor\fontdimen3\font minus
  \fontdimen4\font\relax}
\providecommand{\BIBforeignlanguage}[2]{{%
\expandafter\ifx\csname l@#1\endcsname\relax
\typeout{** WARNING: IEEEtran.bst: No hyphenation pattern has been}%
\typeout{** loaded for the language `#1'. Using the pattern for}%
\typeout{** the default language instead.}%
\else
\language=\csname l@#1\endcsname
\fi
#2}}
\providecommand{\BIBdecl}{\relax}
\BIBdecl

\bibitem{LIME}
\BIBentryALTinterwordspacing
M.~T. Ribeiro, S.~Singh, and C.~Guestrin, ``"why should {I} trust you?":
  Explaining the predictions of any classifier,'' \emph{CoRR}, vol.
  abs/1602.04938, 2016. [Online]. Available:
  \url{http://arxiv.org/abs/1602.04938}
\BIBentrySTDinterwordspacing

\bibitem{DLIME}
\BIBentryALTinterwordspacing
M.~R. Zafar and N.~M. Khan, ``{DLIME:} {A} deterministic local interpretable
  model-agnostic explanations approach for computer-aided diagnosis systems,''
  \emph{CoRR}, vol. abs/1906.10263, 2019. [Online]. Available:
  \url{http://arxiv.org/abs/1906.10263}
\BIBentrySTDinterwordspacing

\bibitem{LIMEtree}
\BIBentryALTinterwordspacing
K.~Sokol and P.~A. Flach, ``Limetree: Interactively customisable explanations
  based on local surrogate multi-output regression trees,'' \emph{CoRR}, vol.
  abs/2005.01427, 2020. [Online]. Available:
  \url{https://arxiv.org/abs/2005.01427}
\BIBentrySTDinterwordspacing

\bibitem{k-LIME}
N.~Gill, M.~Kurka, and W.~Phan, ``Machine learning interpretability with h2o
  driverless ai,'' 2019.

\bibitem{deeplift}
\BIBentryALTinterwordspacing
A.~Shrikumar, P.~Greenside, and A.~Kundaje, ``Learning important features
  through propagating activation differences,'' \emph{CoRR}, vol.
  abs/1704.02685, 2017. [Online]. Available:
  \url{http://arxiv.org/abs/1704.02685}
\BIBentrySTDinterwordspacing

\bibitem{intergratedgradients}
\BIBentryALTinterwordspacing
M.~Sundararajan, A.~Taly, and Q.~Yan, ``Axiomatic attribution for deep
  networks,'' \emph{CoRR}, vol. abs/1703.01365, 2017. [Online]. Available:
  \url{http://arxiv.org/abs/1703.01365}
\BIBentrySTDinterwordspacing

\bibitem{grad-cam}
R.~R. Selvaraju, A.~Das, R.~Vedantam, M.~Cogswell, D.~Parikh, and D.~Batra,
  ``Grad-cam: Why did you say that?'' \emph{arXiv preprint arXiv:1611.07450},
  2016.

\bibitem{survey2018}
\BIBentryALTinterwordspacing
L.~H. Gilpin, D.~Bau, B.~Z. Yuan, A.~Bajwa, M.~A. Specter, and L.~Kagal,
  ``Explaining explanations: An approach to evaluating interpretability of
  machine learning,'' \emph{CoRR}, vol. abs/1806.00069, 2018. [Online].
  Available: \url{http://arxiv.org/abs/1806.00069}
\BIBentrySTDinterwordspacing

\bibitem{survey2019}
A.~Adadi and M.~Berrada, ``Peeking inside the black-box: A survey on
  explainable artificial intelligence (xai),'' \emph{IEEE Access}, vol.~6, pp.
  52\,138--52\,160, 2018.

\bibitem{survey2021}
\BIBentryALTinterwordspacing
P.~Linardatos, V.~Papastefanopoulos, and S.~Kotsiantis, ``Explainable ai: A
  review of machine learning interpretability methods,'' \emph{Entropy},
  vol.~23, no.~1, 2021. [Online]. Available:
  \url{https://www.mdpi.com/1099-4300/23/1/18}
\BIBentrySTDinterwordspacing

\bibitem{surveyNLP2020}
\BIBentryALTinterwordspacing
M.~Danilevsky, K.~Qian, R.~Aharonov, Y.~Katsis, B.~Kawas, and P.~Sen, ``A
  survey of the state of explainable {AI} for natural language processing,''
  \emph{CoRR}, vol. abs/2010.00711, 2020. [Online]. Available:
  \url{https://arxiv.org/abs/2010.00711}
\BIBentrySTDinterwordspacing

\bibitem{madsen2021post}
A.~Madsen, S.~Reddy, and S.~Chandar, ``Post-hoc interpretability for neural
  nlp: A survey,'' \emph{arXiv preprint arXiv:2108.04840}, 2021.

\bibitem{wrongEP_amy}
A.~Rafferty, R.~Nenutil, and A.~Rajan, ``Explainable artificial intelligence
  for breast tumour classification: Helpful or harmful,'' in
  \emph{Interpretability of Machine Intelligence in Medical Image Computing:
  5th International Workshop, iMIMIC 2022, Held in Conjunction with MICCAI
  2022, Singapore, Singapore, September 22, 2022, Proceedings}.\hskip 1em plus
  0.5em minus 0.4em\relax Springer, 2022, pp. 104--123.

\bibitem{wrongEP_2}
N.~Arun, N.~Gaw, P.~Singh, K.~Chang, M.~Aggarwal, B.~Chen \emph{et~al.},
  ``Assessing the (un) trustworthiness of saliency maps for localizing
  abnormalities in medical imaging. arxiv,'' \emph{arXiv preprint
  arXiv:2008.02766}, 2020.

\bibitem{wrongEP_3}
A.~Hedstr{\"o}m, L.~Weber, D.~Bareeva, F.~Motzkus, W.~Samek, S.~Lapuschkin, and
  M.~M.-C. H{\"o}hne, ``Quantus: an explainable ai toolkit for responsible
  evaluation of neural network explanations,'' \emph{arXiv preprint
  arXiv:2202.06861}, 2022.

\bibitem{Lin_eva_gt}
\BIBentryALTinterwordspacing
Y.~Lin, W.~Lee, and Z.~B. Celik, ``What do you see? evaluation of explainable
  artificial intelligence {(XAI)} interpretability through neural backdoors,''
  \emph{CoRR}, vol. abs/2009.10639, 2020. [Online]. Available:
  \url{https://arxiv.org/abs/2009.10639}
\BIBentrySTDinterwordspacing

\bibitem{alignment_eva_1}
J.~Zhang, S.~A. Bargal, Z.~Lin, J.~Brandt, X.~Shen, and S.~Sclaroff, ``Top-down
  neural attention by excitation backprop,'' \emph{International Journal of
  Computer Vision}, vol. 126, no.~10, pp. 1084--1102, 2018.

\bibitem{alignment_eva_2}
M.~Yang and B.~Kim, ``Benchmarking attribution methods with relative feature
  importance,'' \emph{arXiv preprint arXiv:1907.09701}, 2019.

\bibitem{alignment_eva_3}
Y.~Zhou, S.~Booth, M.~T. Ribeiro, and J.~Shah, ``Do feature attribution methods
  correctly attribute features?'' in \emph{Proceedings of the AAAI Conference
  on Artificial Intelligence}, vol.~36, no.~9, 2022, pp. 9623--9633.

\bibitem{eva_gt_yang}
\BIBentryALTinterwordspacing
M.~Yang and B.~Kim, ``{BIM:} towards quantitative evaluation of
  interpretability methods with ground truth,'' \emph{CoRR}, vol.
  abs/1907.09701, 2019. [Online]. Available:
  \url{http://arxiv.org/abs/1907.09701}
\BIBentrySTDinterwordspacing

\bibitem{eva_gt_Arras}
L.~Arras, A.~Osman, and W.~Samek, ``Clevr-xai: a benchmark dataset for the
  ground truth evaluation of neural network explanations,'' \emph{Information
  Fusion}, vol.~81, pp. 14--40, 2022.

\bibitem{eva_gt_Adebayo}
\BIBentryALTinterwordspacing
J.~Adebayo, M.~Muelly, I.~Liccardi, and B.~Kim, ``Debugging tests for model
  explanations,'' 2020. [Online]. Available:
  \url{https://arxiv.org/abs/2011.05429}
\BIBentrySTDinterwordspacing

\bibitem{eva_gt_Holzinger}
A.~Holzinger, A.~Carrington, and H.~M{\"u}ller, ``Measuring the quality of
  explanations: the system causability scale (scs) comparing human and machine
  explanations,'' \emph{KI-K{\"u}nstliche Intelligenz}, vol.~34, no.~2, pp.
  193--198, 2020.

\bibitem{hooker_performance_eva_7}
S.~Hooker, D.~Erhan, P.-J. Kindermans, and B.~Kim, ``A benchmark for
  interpretability methods in deep neural networks,'' \emph{Advances in neural
  information processing systems}, vol.~32, 2019.

\bibitem{petsiuk_performance_eva_3}
V.~Petsiuk, A.~Das, and K.~Saenko, ``Rise: Randomized input sampling for
  explanation of black-box models,'' \emph{arXiv preprint arXiv:1806.07421},
  2018.

\bibitem{Samek_eva_performance}
\BIBentryALTinterwordspacing
W.~Samek, A.~Binder, G.~Montavon, S.~Bach, and K.~M{\"{u}}ller, ``Evaluating
  the visualization of what a deep neural network has learned,'' \emph{CoRR},
  vol. abs/1509.06321, 2015. [Online]. Available:
  \url{http://arxiv.org/abs/1509.06321}
\BIBentrySTDinterwordspacing

\bibitem{yeh_performance_eva_6}
C.-K. Yeh, C.-Y. Hsieh, A.~Suggala, D.~I. Inouye, and P.~K. Ravikumar, ``On the
  (in) fidelity and sensitivity of explanations,'' \emph{Advances in Neural
  Information Processing Systems}, vol.~32, 2019.

\bibitem{eva_performance_ismail}
\BIBentryALTinterwordspacing
A.~A. Ismail, M.~K. Gunady, H.~C. Bravo, and S.~Feizi, ``Benchmarking deep
  learning interpretability in time series predictions,'' \emph{CoRR}, vol.
  abs/2010.13924, 2020. [Online]. Available:
  \url{https://arxiv.org/abs/2010.13924}
\BIBentrySTDinterwordspacing

\bibitem{arras2019eva_performance}
L.~Arras, A.~Osman, K.-R. M{\"u}ller, and W.~Samek, ``Evaluating recurrent
  neural network explanations,'' \emph{arXiv preprint arXiv:1904.11829}, 2019.

\bibitem{lin_performance_eva_4}
Z.~Q. Lin, M.~J. Shafiee, S.~Bochkarev, M.~S. Jules, X.~Y. Wang, and A.~Wong,
  ``Do explanations reflect decisions? a machine-centric strategy to quantify
  the performance of explainability algorithms,'' \emph{arXiv preprint
  arXiv:1910.07387}, 2019.

\bibitem{chen_performance_eva_2}
J.~Chen, L.~Song, M.~J. Wainwright, and M.~I. Jordan, ``L-shapley and
  c-shapley: Efficient model interpretation for structured data,'' \emph{arXiv
  preprint arXiv:1808.02610}, 2018.

\bibitem{ASR_EP}
\BIBentryALTinterwordspacing
X.~Wu, P.~Bell, and A.~Rajan, ``Explanations for automatic speech
  recognition,'' 2023. [Online]. Available:
  \url{https://arxiv.org/abs/2302.14062}
\BIBentrySTDinterwordspacing

\bibitem{TIMIT}
J.~S. Garofolo, L.~F. Lamel, W.~M. Fisher, J.~G. Fiscus, and D.~S. Pallett,
  ``Darpa timit acoustic-phonetic continous speech corpus cd-rom. nist speech
  disc 1-1.1,'' \emph{NASA STI/Recon technical report n}, vol.~93, p. 27403,
  1993.

\bibitem{kaldi}
D.~Povey, A.~Ghoshal, G.~Boulianne, L.~Burget, O.~Glembek, N.~Goel,
  M.~Hannemann, P.~Motlicek, Y.~Qian, P.~Schwarz \emph{et~al.}, ``The kaldi
  speech recognition toolkit,'' in \emph{IEEE 2011 workshop on automatic speech
  recognition and understanding}, no. CONF.\hskip 1em plus 0.5em minus
  0.4em\relax IEEE Signal Processing Society, 2011.

\bibitem{commonvoice:2020}
R.~Ardila, M.~Branson, K.~Davis, M.~Henretty, M.~Kohler, J.~Meyer, R.~Morais,
  L.~Saunders, F.~M. Tyers, and G.~Weber, ``Common voice: A
  massively-multilingual speech corpus,'' in \emph{Proceedings of the 12th
  Conference on Language Resources and Evaluation (LREC 2020)}, 2020, pp.
  4211--4215.

\bibitem{tukey}
J.~W. Tukey \emph{et~al.}, \emph{Exploratory data analysis}.\hskip 1em plus
  0.5em minus 0.4em\relax Reading, MA, 1977, vol.~2.

\end{thebibliography}

\end{document}